\documentclass[letterpaper, 10 pt, conference]{ieeeconf}  

\IEEEoverridecommandlockouts                              

\usepackage{amsmath} 
\usepackage{amssymb}  
\usepackage{booktabs}
\usepackage{algorithm}
\usepackage{algpseudocode}
\usepackage{booktabs}
\usepackage{graphicx}
\usepackage{subcaption}
\usepackage{xcolor}
\usepackage{cuted}
\usepackage{caption}
\usepackage{changes}
\usepackage[table]{xcolor}

\algnewcommand\algorithmicinput{\textbf{Input:}}
\algnewcommand\algorithmicoutput{\textbf{Output:}}
\algnewcommand\Input{\item[\algorithmicinput]}
\algnewcommand\Output{\item[\algorithmicoutput]}

\definechangesauthor[name={Ankit}, color=blue]{A}
\definecolor{sehoonpink}{RGB}{214,51,132}
\setdeletedmarkup{}   
\renewcommand{\deleted}[2][]{}   
\definechangesauthor[name={Sehoon}, color=sehoonpink]{S}

\title{\LARGE \bf
Rapid co-design of Buoyancy-assisted robots for Challenging Locomotion using Gaussian Evolutionary Specialists 
}

\author{Ankit Sinha$^{1}$, Nitish Sontakke$^{1}$, Dennis Hong$^{2}$, Yusuke Tanaka$^{3}$ and Sehoon Ha$^{1}$ 
\thanks{$^{1}$Ankit, Nitish and Sehoon are with the School of Interactive Computing, Georgia Institute of Technology, TSRB 85 5th St NW, Atlanta, GA - 30332, USA
      {\tt\small \{asinha389,nitishsontakke,sha9\}@gatech.edu}}%
\thanks{$^{2}$ Dennis is with the Dept. of Mechanical and Aerospace Engineering, University of California, Los Angeles, 420 Westwood Plaza, CA 90095, USA
        {\tt\small dennishong@ucla.edu}}%
\thanks{$^{3}$ Yusuke is with the Robotic Systems Lab, ETH Zurich, Leonhardstrasse 21, 8092 Zurich, Switzerland
        {\tt\small yutanaka@ethz.ch}}
}

\begin{document}
\maketitle
\thispagestyle{empty}
\pagestyle{empty}
\bstctlcite{IEEEexample:BSTcontrol}

\begin{abstract}
Designing high-performance legged robots requires jointly optimizing morphology and control. 
Model-free reinforcement learning (RL) enables learning complex controllers during co-design, but repeatedly training policies across candidate morphologies makes design optimization computationally expensive. Existing approaches can amortize policy learning across morphologies, but a single shared policy tends to converge toward one control strategy regardless of morphology, which we refer to as \textit{behavioral diversity collapse}.
We introduce Gaussian Evolutionary Specialists (GES), an RL-based co-design framework that maintains a population of specialist policies whose design-space territories evolve according to their relative control performance. We evaluate GES on the Buoyancy-Assisted Light Legged Unit (BALLU) robot for obstacle traversal and ramp walking over design spaces of up to four dimensions. 
GES achieves higher best design performance than Monolithic and Mixture-of-Experts baselines, with gains of \(5.0-9.0~cm\) in obstacle height and \(10.0-12.7\) points in ramp grade. For the same 200 design-optimization evaluations, GES reduces wall-clock co-design time by 67.4\% 
compared with retraining a policy for every candidate. Finally, teleoperated hardware experiments evaluate the optimized BALLU morphology, which traverses obstacles of up to 24~cm, compared with 8~cm on the original design.

\end{abstract}

\section{Introduction}
\label{sec:intro}

Robot co-design jointly optimizes morphology and control to achieve task performance that
neither component can reach in isolation~\cite{gupta2021embodied,bjelonic2023learning}.
For instance, optimizing link lengths in a robotic arm expands the reachable workspace for a
target task, while tuning leg proportions in a legged robot shifts the trade-off between
energy efficiency and agility~\cite{bjelonic2023learning}.
The Buoyancy-Assisted Light Legged Unit (BALLU)~\cite{chae2021ballu2,hong2025buoyant}
studied in this work is one such platform: its sensitivity to buoyancy, leg proportions,
and passive joint properties makes morphology integral to its locomotion capability.
Moreover, BALLU's lightweight structure, spring--servo actuation, and modeling inaccuracies
make multi-parameter morphology--control optimization challenging.

\begin{figure}[t]
    \centering
    \includegraphics[width=1.0\linewidth]{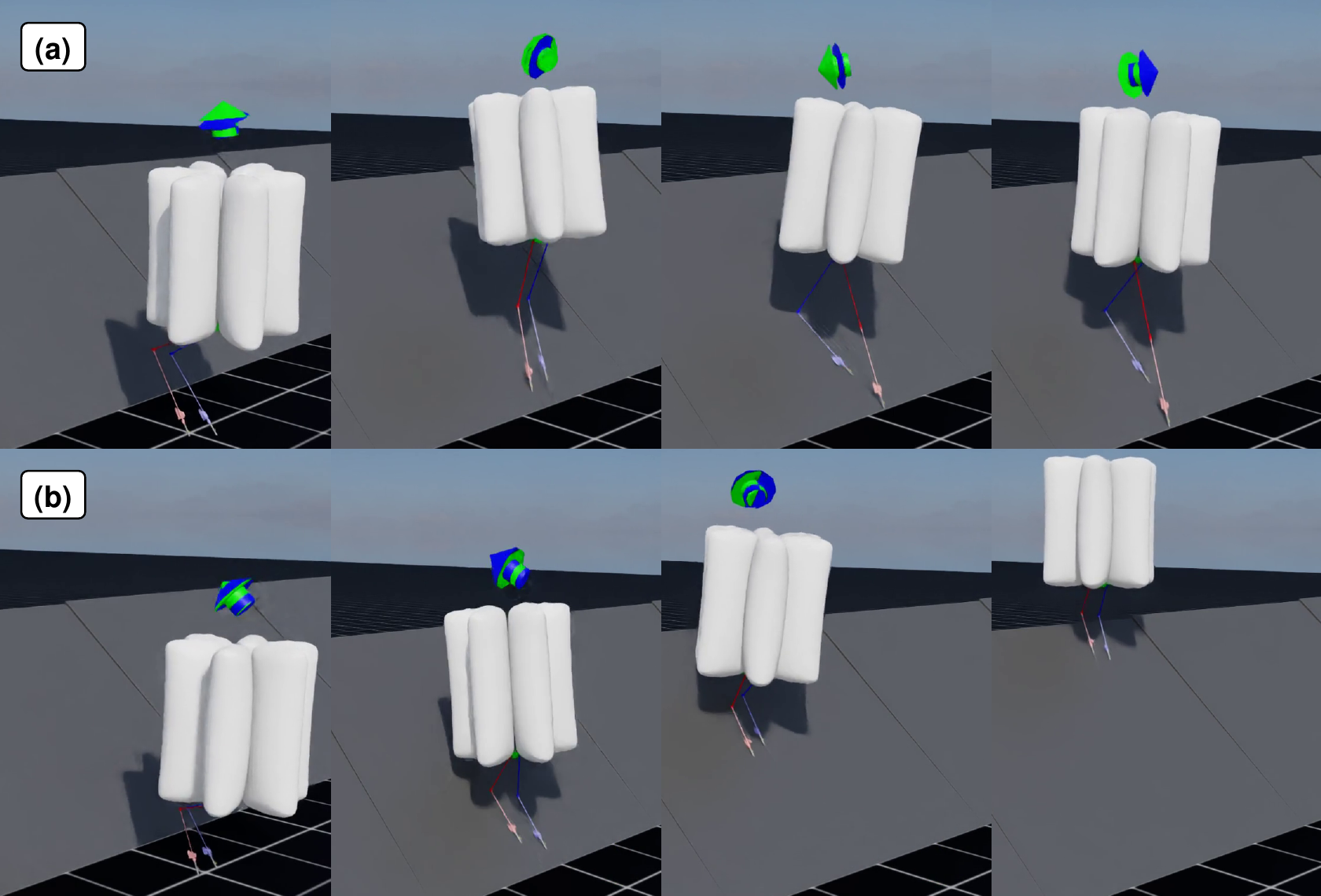}
    \caption{The BALLU robot facing a ramp of slope grade 80\%\protect{ (\(\approx 38.7^\circ\))} -- (a) baseline design with long limbs is unable to walk up. (b) \textit{optimized} design with shorter limbs is more athletic, progresses steadily and reaches the top of the ramp.} 
    \label{fig:sim-motion-strip}
    \vspace{-2em}
\end{figure}

Model-based co-design methods can efficiently couple morphology optimization with
trajectory or controller optimization~\cite{ha2018coopt,chadwick2020vitruvio,
fadini2024codesigning}. Model-free reinforcement learning (RL) provides an alternative
for learning controllers during co-design, particularly when complex behaviors are difficult
to prescribe through a model-based controller. However, a straightforward RL-based
co-design pipeline requires policy optimization for each candidate morphology, making
large-scale design search computationally expensive~\cite{bjelonic2023learning}.

Recent learning-based approaches mitigate this cost by sharing policies, value functions,
or experience across morphologies~\cite{Schaff2018JointlyLT,Luck2019DataefficientCOA,
bjelonic2023learning}. In particular, a morphology-conditioned policy can amortize
controller learning across the design space and evaluate new designs without training a
policy from scratch. However, a single shared policy must represent effective control
strategies across potentially heterogeneous morphologies. In our experiments with BALLU,
we find that a monolithic design-conditioned policy tends toward similar locomotion
behaviors across different designs, limiting its performance over the design space. We
refer to this observed phenomenon as \textit{behavioral diversity collapse}.

A natural approach to increasing specialization is a Mixture-of-Experts architecture (MoE)~\cite{jacobs1991adaptive}, where multiple experts can represent different control
strategies. Yet, end-to-end MoE training does not guarantee such specialization. In our
experiments, the learned router assigns similar design distributions across experts,
resulting in similar behaviors. This motivates explicitly coupling policy specialization
with morphology-space partitioning rather than relying on end-to-end routing alone.

We propose \textit{Gaussian Evolutionary Specialists} (GES), an RL-based
morphology--control co-design framework that amortizes controller learning across the
design space through an evolving population of specialist policies. Each specialist is
associated with a Gaussian territory in morphology space and is trained on designs sampled
from that region. Through iterative training, competitive probing, and territory updates,
specialists adapt toward regions where their control strategies perform well. The resulting
specialist population is then reused to evaluate candidate morphologies during design
optimization, avoiding policy training from scratch for every candidate.

We validate GES on the BALLU~\cite{chae2021ballu2, sontakke2023residual} platform for hard locomotion tasks that make co-design both necessary and challenging. Our main contributions are:
\begin{itemize}
    \item We introduce GES, an RL-based morphology--control co-design framework that amortizes controller learning across designs using an evolving population of morphology-specialized policies.


    \item 
    We characterize \textit{behavioral diversity collapse} in RL-based design optimization, and provide a quantitative analysis showing that GES mitigates it.

    \item 
    We demonstrate on the BALLU platform that GES discovers higher-performing designs than universal-policy baselines while reducing co-design wall-clock time by 67.4\% relative to per-design policy training.
    
\end{itemize}
\section{RELATED WORK}

\noindent\textbf{Robot Controller Co-Design.}
Robot co-design has been approached through differentiable optimization~\cite{thomaszewski2014computational, bharaj2015computational, ha2017joint, ha2018coopt, xu2021differentiable}, graph and grammar-based morphology search~\cite{zhao2020robogrammar,xu2021moghs,hu2022glso}, evolutionary methods~\cite{gupta2021embodied}, and reinforcement learning~\cite{Schaff2018JointlyLT,Luck2019DataefficientCOA}. Model-based approaches can efficiently couple hardware design with trajectory or controller optimization. For example, \textit{Vitruvio} combines trajectory optimization with evolutionary \deleted[id=S]{design }search~\cite{chadwick2020vitruvio}, while Fadini et al.~\cite{fadini2024codesigning} formulate bi-level hardware--motion optimization for dynamic quadrupedal behaviors. Related methods jointly optimize humanoid hardware and motion for physical interaction~\cite{sartore2022optimization,sartore2023codesign}. In contrast, model-free RL enables learning complex controllers during co-design, but repetitive training for each candidate incurs substantial computational cost.

\noindent\textbf{Learning-based co-design across morphologies.}
Prior methods reduce this cost by sharing information or learned representations across designs. 
Schaff \textit{et al.}~\cite{Schaff2018JointlyLT} jointly learn a design-conditioned controller and morphology distribution, while Luck \textit{et al.}~\cite{Luck2019DataefficientCOA} reuse prior morphology--behavior evaluations to improve data efficiency. 
\textit{Hardware-as-a-Policy}~\cite{chen2021hardware}, \textit{Transform2Act}~\cite{yuan2022transform2act}, and \textit{BodyGen}~\cite{lu2025bodygen} further integrate morphology optimization with policy learning. 
Dai \textit{et al.} \cite{dai2026efficient} formulate co-design as a Stackelberg game and explicitly account for controller adaptation during morphology optimization. More closely related to our setting, Bjelonic et al.~\cite{bjelonic2023learning} train a morphology-conditioned policy and reuse it to evaluate designs during Bayesian optimization; \textit{Shape Your Body} \cite{bohlinger2026shape} similarly reuses a learned value function for design optimization, while \textit{House of Dextra} \cite{fay2026house} employs cross-embodied control for scalable hand co-design. Strgar and Kriegman~\cite{strgar2026accelerated} pre-train controllers across designs and identify population diversity collapse during joint morphology--control optimization. GES similarly amortizes policy learning across designs, but uses evolving specialist policies to retain distinct control strategies across the morphology space.

\noindent\textbf{Quality-diversity methods.}
Quality-Diversity (QD) methods such as MAP-Elites~\cite{mouret2015illuminating} maintain diverse, high-performing solutions across predefined behavioral or design descriptors. Such methods have been applied to morphology--control co-evolution~\cite{nordmoen2021mapelites,zardini2021seeking}, including separate archives for morphological and controller diversity. Unlike QD, our objective is not to produce a diverse repertoire of final robot designs. GES instead maintains diversity among specialist \deleted[id=S]{control }policies used to support the search for high-performing designs.

\noindent\textbf{Universal policies and specialist learning.}
Universal controllers share control knowledge across robot embodiments using morphology-conditioned policies~\cite{feng2023genloco,sorokin2023on}, modular architectures, or cross-embodiment representations. However, learning across heterogeneous morphologies can introduce competing objectives, related to gradient interference observed in multi-task learning~\cite{yu2020gradient,Liu2021ConflictAverseGD}. Mixture-of-Experts (MoE) architectures~\cite{jacobs1991adaptive} offer an alternative through learned specialization, but end-to-end routing can itself suffer from poorly differentiated experts or representation collapse~\cite{chi2022repcollapseMoE}. GES instead explicitly initializes localized specialists and evolves their morphology-space regions according to relative control performance.

\section{PROBLEM FORMULATION}
\label{sec:problem}


\begin{figure}[t]
    \centering
    \includegraphics[width=0.42\linewidth]{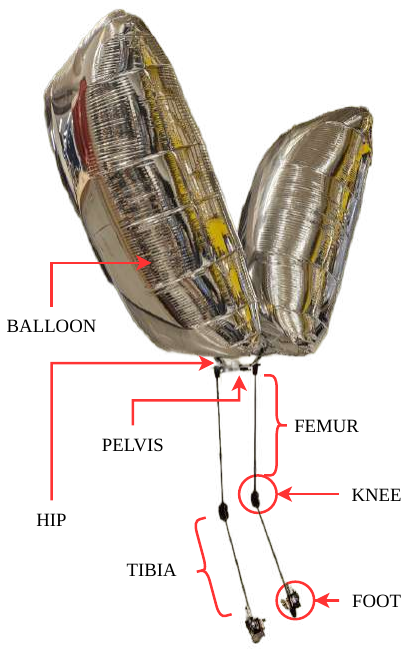}
    \caption{BALLU Hardware}
    \label{fig:ballu-figure}
    \vspace{-2em}
\end{figure}
\subsection{BALLU Platform}

BALLU~\cite{chae2021ballu2, hong2025buoyant} is a bipedal robot that uses helium-filled
balloons to offset the majority of its weight. 
The platform (Fig. \ref{fig:ballu-figure}) consists of five rigid links: a pelvis and two upper and
lower limbs connected by spring-loaded knee joints actuated by a
single servo per leg, which gives the robot two actuated degrees of freedom
in total.
Its lightweight construction and fall-safe operation make it well-suited for human-facing applications like environmental monitoring and social robotics.

Despite its mechanical simplicity, BALLU presents significant control challenges. 
The balloons introduce severe aerodynamic nonlinearities: buoyancy and drag forces that are difficult to model analytically. 
Moreover, aerodynamic effects vary with the balloon's helium fill level and environmental conditions~\cite{chae2021ballu2}. 
These dynamics make model-free RL an attractive approach to controller design, as demonstrated by Sontakke et al.~\cite{sontakke2023residual}. 

\subsection{Design Space}
\label{sec:design-space}
The BALLU morphology design space, $\mathcal{D}$, is parameterized by four variables: femur length $l_f$, tibia length $l_t$, gravity compensation ratio (\(\mathrm{GCR}=\frac{\text{buoyancy force}}{\text{robot weight}}\)), and spring stiffness, $\mathrm{SPCF}$.
At GCR=1.0, the robot is effectively weightless; lower values increase the effective gravitational load on the legs.
The $\mathrm{SPCF}$ governs knee joint compliance, which directly affects energy storage during dynamic maneuvers.
The parameter bounds and the baseline BALLU design \cite{chae2021ballu2} are summarized in Tab.~\ref{tab:design_space}.
\begin{table}[t]
  \centering
  \caption{BALLU design parameters, initial and optimized designs for each task: Obstacle Traversal and Ramp Locomotion. \protect{GCR: gravity comp.\ ratio; SPCF: spring coeff.}}
  \label{tab:design_space}
  \begin{tabular}{llllll}
    \toprule
    Parameter  & \shortstack{Lower\\bound} & \shortstack{Upper\\bound} & \shortstack{Baseline\\Design} & \shortstack{Final\\Design\\Obstacle} & \shortstack{Final\\Design\\Ramp}\\
    \midrule
    Femur (mm)        & 200 & 500 & 450 & 458 & 300 \\
    Tibia (mm)        & 200 & 500 & 450 & 458 & 280 \\
    GCR   & 0.65    & 0.90    & 0.70    & 0.88   & 0.70\\
    SPCF (Nm/rad) & \(10^{-3}\)  & \(10^{-2}\)   & \(2\cdot 10^{-3}\) & \(8\cdot 10^{-3}\) & \(5\cdot 10^{-3}\) \\
    \bottomrule
  \end{tabular}
  \vspace{-1em}
\end{table}
We conduct experiments across three design parameter spaces: $\mathcal{D}^{2}=\{(\mathrm{GCR},\mathrm{SPCF}) : l_f=\bar{l}_f,\; l_t=\bar{l}_t\}\subset \mathbb{R}^{2}$, $\mathcal{D}^{3}=\{(\mathrm{GCR},\mathrm{SPCF},l) : l_f=l_t=l\}\subset \mathbb{R}^{3}$ and $\mathcal{D}^{4}=\{(\mathrm{GCR},\mathrm{SPCF},l_f, l_t) \}\subset \mathbb{R}^{4}$.

\subsection{Bi-level Co-design Objective}
The goal of robot co-design is to jointly find a morphology
$\boldsymbol{\theta}_m$ and a control policy
$\boldsymbol{\theta}_c$ that maximize task performance:
\begin{equation}
  \max_{\boldsymbol{\theta}_m,\, \boldsymbol{\theta}_c}
    \; J(\boldsymbol{\theta}_c,\, \boldsymbol{\theta}_m).
  \label{eq:joint}
\end{equation}
This joint optimization is intractable to solve directly due to the
interdependency between morphology and control. We therefore decompose
it into a bi-level form:
\begin{equation}
  \max_{\boldsymbol{\theta}_m}
    \; J\!\left(\boldsymbol{\theta}_c^*,\,
               \boldsymbol{\theta}_m\right),
  \;
  \boldsymbol{\theta}_c^*
    = \arg\max_{\boldsymbol{\theta}_c}
        \; J(\boldsymbol{\theta}_c,\, \boldsymbol{\theta}_m),
  \label{eq:bilevel}
\end{equation}
where the inner loop finds the optimal controller for a given morphology
and the outer loop searches over the morphology space to
maximize the resulting performance. 


\subsection{MDP Formulation}
\label{sec:mdp}

We model locomotion control as a Markov Decision Process (MDP) defined
by the tuple $(\mathcal{S}, \mathcal{A}, \mathcal{T}, \mathcal{R},
\gamma)$.
The proprioceptive state $s_t \in \mathcal{S}$
comprises the robot's base linear and angular velocity, joint positions,
joint velocities, and the last action.
The augmented state $\tilde{s}_t = [s_t,\, \boldsymbol{\theta}_m]$ appends the design vector. 
The action $a_t \in \mathbb{R}^2$ specifies joint
position targets for the left and right knees, which are tracked by a
PD controller that converts targets to joint torques.
The policy $\pi_{\boldsymbol{\theta}_c}(a_t \mid \tilde{s_t})$
is parameterized as a Gaussian MLP and optimized via PPO to maximize
expected discounted return:
\begin{equation}
  \max_{\boldsymbol{\theta}_c} \; \mathbb{E}\!\left[
    \sum_{t=0}^{T} \gamma^t \, r(s_t, a_t)
  \right]
  \label{eq:rl_objective}
\end{equation}



\noindent\textbf{Reward functions.} We have two different tasks described in Sec.~\ref{sec:exp-setup}. For obstacle traversal, the reward combines a point-goal navigation term, a forward velocity term, and a jumping term that encourages toe clearance:
\begin{equation}
  r_{\mathrm{obs}} =
    w_1\underbrace{(-\|g - s\|_2^2)}_{r_{\mathrm{nav}}}
    + w_2\underbrace{(v_x)}_{r_{\mathrm{vel}}}
    + w_3\underbrace{\left(
        e^{c \cdot z_{\mathrm{toe}}} - 1
      \right)}_{r_{\mathrm{jump}}},
  \label{eq:reward_obs}
\end{equation}
where $g$ is the goal position, $s$ is the robot's current base
position, $v_x$ is the forward velocity, and
$z_{\mathrm{toe}} = \min(z_{\mathrm{toe}}^L, z_{\mathrm{toe}}^R)$ is the minimum toe height across both legs.
For ramp walking, the jumping term is dropped as we are interested in a smooth and stable walking gait:
\(
  r_{\mathrm{ramp}} = w_1 (-\|g - s\|_2^2) + w_2 \, v_x.
\)
We set $w_1=5, w_2=3, w_3=1$ and $c=1.73$. Each episode lasts 20~s.

\section{GAUSSIAN EVOLUTIONARY SPECIALISTS}
\label{sec:ges}

\subsection{Motivation for Specialist Policies}

Amortizing policy learning across morphologies avoids training a controller from scratch for every design candidate. However, a single morphology-conditioned policy must represent effective control strategies across the entire design space. In our experiments, such a monolithic policy converges toward similar locomotion behaviors across different morphologies, which we refer to as \textit{behavioral diversity collapse}. A multi-headed policy like end-to-end Mixture-of-Experts (MoE) provides additional capacity, but does not guarantee specialization; we observe nearly uniform routing across experts, consistent with representation collapse~\cite{chi2022repcollapseMoE}. These observations motivate GES, which explicitly partitions the morphology space among specialist policies and evolves these territories according to relative control performance.

\begin{figure}[t]
    \centering
    \includegraphics[width=0.42\textwidth]{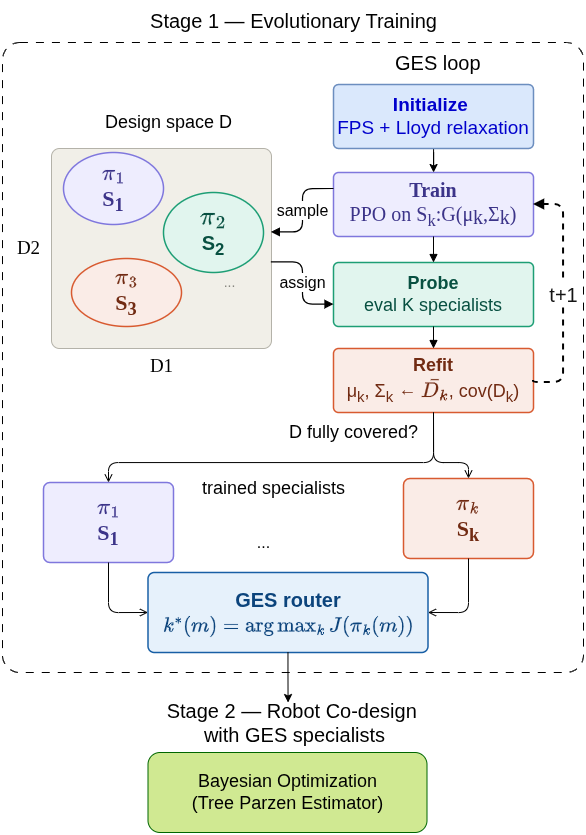}
    \caption{GES: Our proposed algorithm for training a mixture of specialists through design space evolution. Sub-stage 1 (\textbf{Initialize}) consists of spawning Gaussian centers which are dispersed well enough to cover the design space. Sub-stage 2 (\textbf{Train}) consists of training each specialist policy by sampling designs from its associated Gaussian territory. In sub-stage 3 (\textbf{Probe}), we sample designs from the Gaussian territory edges and assign the designs through competition \textit{i.e.} a design goes to the highest performing policy. In sub-stage 4 (\textbf{Refit}), each specialist's Gaussian is updated to reflect the acquired territories. This process repeats until MC coverage (Sec. \ref{sec:ges_algo}) of the design space reaches 95\%.} 
    \label{fig:GES_BO}
    \vspace{-2em}
\end{figure}

\subsection{GES Algorithm}
\label{sec:ges_algo}

The goal of GES is to approximate $J(\boldsymbol{\theta}_c^*,\, \boldsymbol{\theta}_m)$ (Eq.~\eqref{eq:bilevel}) across all $\boldsymbol{\theta}_m \in \mathcal{D}$, without requiring per-design RL policy training.
Fig.~\ref{fig:GES_BO} illustrates the full pipeline.
The key insight behind GES is to \emph{decouple} design-space
partitioning from expert specialization.
Rather than learning a router end-to-end alongside the experts, GES
uses a geometrically defined partition that evolves iteratively based on
each specialist's empirical performance.

Let \(\mathcal{D}\) be the design space of dimension \(d\). Formally, GES maintains $k \in \{1,\ldots,K\}$ specialists each
represented by a Gaussian territory with mean $\boldsymbol{\mu}^{(k)} \in \mathbb{R}^d$ and variances
$(\sigma^{2}_{11})^{(k)}, \ldots, (\sigma^{2}_{dd})^{(k)}$.
This territory defines the distribution from which training designs are sampled for specialist~$\pi_k$, as well as the region of~$\mathcal{D}$ that specialist is responsible for at BO query time. GES proceeds in three repeating phases until the territories
collectively cover~$\mathcal{D}$:

\subsubsection{Initialization}. We initialize $K$ Gaussian centers in the normalized design space $\mathcal{X} = [0,1]^d$. We seek $K$ centers $\{\mu_k\}_{k=1}^K \subset \mathcal{X}$ that provide good coverage of $\mathcal{X}$, which can be formulated as the following space-filling objective:
\begin{equation}
\min_{\{\mu_k\}} \; \max_{x \in \mathcal{X}} \; \min_{1 \le k \le K} \|x - \mu_k\|_2.
\end{equation}

Directly solving this problem is intractable, so we adopt a two-stage approximation that combines dispersion and coverage refinement.

\noindent\textbf{Farthest Point Sampling (FPS).}
Starting from a random initial center, we iteratively select the candidate from a uniform sample set $\mathcal{C} \subset \mathcal{X}$ that is farthest from all existing centers. This yields a well-dispersed set of $K$ centers.

\noindent\textbf{Lloyd Relaxation.}
We then refine the centers via $T$ iterations of Lloyd's algorithm~\cite{lloyd1982least}, which iteratively reassigns each center to the centroid of its Voronoi region. After $T$ iterations, we obtain centers that are both well-separated and approximately space-filling.

\subsubsection{Iterative Territory Evolution}

\noindent\textbf{Train.} Each specialist~$\pi_k$ samples $n_{\mathrm{train}}$
designs from its current Gaussian territory
$\mathcal{G}(\boldsymbol{\mu}_k, \boldsymbol{\Sigma}_k)$ and trains
for $n_{\mathrm{ppo}}$ PPO steps, and resumes from its previous checkpoint to retain learned behaviors.

\noindent\textbf{Probe.} $F$ candidate designs are sampled near the $2\sigma$
borders of all Gaussian territories. Each candidate is evaluated by all
$K$ specialists, and assigned to the specialist that achieves the
highest task performance score $J$ on that design:
\begin{equation}
  k^* = \arg\max_{k} \; J(\pi_k, c), \quad
  \eta_{k^*} \leftarrow \eta_{k^*} \cup \{c\}.
  \label{eq:probe}
\end{equation}
Here, \(\eta_k\) implies all designs assigned to specialist \(k\). This competitive probing at territory borders drives specialists to
expand into regions where they outperform other specialists, which yields naturally performance-grounded boundaries.

\noindent\textbf{Refit.} Each specialist's Gaussian territory is refit to the
set of designs it has won:
\begin{equation}
  \boldsymbol{\mu}_k \leftarrow \text{mean}(\eta_k),\:
  \boldsymbol{\Sigma}_k \leftarrow \text{diagcov}(\eta_k),\;
   \forall k \in \{1...K\}
  \label{eq:refit}
\end{equation}
The diagonal covariance constraint keeps each territory axis-aligned
and interpretable in the design parameter space.
As iterations progress, territories expand from their initially small,
localized regions to collectively tile~$\mathcal{D}$, and each
specialist concentrates its expertise in the morphological regime where it naturally excels. During iterative territory evolution, we measure coverage of the design space via Monte-Carlo (MC) sampling as defined next.

\noindent\textbf{MC coverage}. 
For a design $\boldsymbol{\theta} = (\theta_1,\ldots,\theta_{\|D\|})^\top \in \mathcal{D}$,
let us define the un-normalized multivariate density contribution of specialist $k$ as
\begin{equation}
f_k(\boldsymbol{\theta}) =
\exp\bigl(-\sum_{d=1}^{\|D\|}
(\theta_d - \mu^{(k)}_{d})^{2}\,/\,\bigl(2 (\sigma^{2}_{dd})^{(k)}\bigr)\bigr)
\end{equation}
We declare $\boldsymbol{\theta}$ \emph{covered} by the mixture if
$\max_{1 \le k \le K} f_k(\boldsymbol{\theta}) > \tau$ (\(=e^{-2} \approx 0.135\)) corresponding to the two-standard-deviation
contour under a diagonal Gaussian in the implementation.
The \emph{Monte Carlo coverage} is
$c = \mathbb{P}_{\boldsymbol{\theta} \sim \mathrm{Unif}(\mathcal{D})}
\bigl[\max_{k} f_k(\boldsymbol{\theta}) > \tau\bigr]$,
estimated by drawing $N_{\mathrm{mc}}$ i.i.d.\ samples
$\boldsymbol{\theta}^{(1)},\ldots,\boldsymbol{\theta}^{(N_{\mathrm{mc}})}$
from $\mathrm{Unif}(\mathcal{D})$:
\begin{equation}
  \hat{c} \;=\; \frac{1}{N_{\mathrm{mc}}}
  \sum_{i=1}^{N_{\mathrm{mc}}}
  \mathbf{1}\!\left\{ \max_{k} f_k\!\left(\boldsymbol{\theta}^{(i)}\right) > \tau \right\}.
\end{equation}
The \texttt{Train-Probe-Refit} iteration stops when the MC coverage (line 18 Alg. \ref{alg:ges}) of the design space reaches \(95\%\) under \(N_{mc}=10^4\). We set initial variance \(\sigma_0\) such that initial coverage \(\hat{c}_0 = 70 \%\) and \(F=1000\).

\begin{algorithm}[!t]
  \caption{Gaussian Evolutionary Specialists (GES)}
  \label{alg:ges}
  \begin{algorithmic}[1]

    \Input Design space $\mathcal{D}$, \# specialists
      $K$, \# probe designs $F$,
      train designs $n_{\mathrm{train}}$,
      initial var $\sigma_0^2$,
      PPO steps $n_{\mathrm{ppo}}$
    \Output $K$ specialists
      $\{\pi_k\}_{k=1}^{K}$ covering $\{(\boldsymbol{\mu}_k, \boldsymbol{\Sigma}_k)\}_{k=1}^{K}$

    \Statex \textit{// Initialization}
    \State $\{\boldsymbol{\mu}\}_{k=1}^K \gets \text{FPSwithLloydRelax}(\mathcal{D}, K)$
    \For{$k \gets 1$ \textbf{to} $K$}
      
      \State $\pi_k, \eta_k, \Sigma_k \gets$ random init MLP, $\emptyset, \sigma^2_0 I$
    \EndFor

    \State $\mathrm{covered} \gets \mathbf{false}$; $t \gets 0$
    \While{$\mathrm{covered} = \mathbf{false}$}
      \Statex \textit{// Train}
      \For{$k \gets 1$ \textbf{to} $K$}
        \State $\mathcal{B}_k \gets
          \mathrm{Sample}\!\left(
            n_{\mathrm{train}},\,
            \mathcal{G}(\boldsymbol{\mu}_k,
            \boldsymbol{\Sigma}_k)\right)$
        \State $\pi_k \gets
          \mathrm{PPO}\!\left(
            \pi_k^{t-1},\,
            \mathcal{B}_k,\,
            n_{\mathrm{ppo}}\right)$
      \EndFor

      \Statex \textit{// Probe}
    \State $\mathcal{C} \gets \mathrm{SampleBorder}(F, \{(\boldsymbol{\mu}_k, \boldsymbol{\Sigma}_k)\}_{k=1}^{K})$
    \ForAll{$c \in \mathcal{C}$}
        \State $k^* \gets \arg\max_{k} \; J(\pi_k, c)$
        \State $\eta_{k^*} \gets \eta_{k^*} \cup \{c\}$
    \EndFor

      \Statex \textit{// Refit}
      \State $\boldsymbol{\mu_k}, \boldsymbol{\Sigma_k} \gets \mathrm{mean}(\eta_k), \mathrm{diagcov}(\eta_k) \forall k = \{1\cdots K\}$ 

      \State $\mathrm{covered} \gets
        \mathrm{CheckCoverage}\!\left(
          \{(\boldsymbol{\mu}_k,
          \boldsymbol{\Sigma}_k)\}_{k=1}^{K},\,
          \mathcal{D}\right)$

        \State $t \gets t + 1$

    \EndWhile

    \State \Return
      $\{\pi_k,\,
      \boldsymbol{\mu}_k,\,
      \boldsymbol{\Sigma}_k\}_{k=1}^{K}$

  \end{algorithmic}
\end{algorithm}

\section{RESULTS}
\label{sec:results}


\subsection{Experimental Setup}
\label{sec:exp-setup}

\noindent\textbf{Simulation Model.}
We model BALLU as a rigid-body articulation in Isaac Lab 2.0
\cite{mittal2025isaaclab} and Isaac Sim 4.5, with simulation and control
frequencies of 200 and 20~Hz, respectively. Buoyancy is applied to the
balloon body as an external force
($F_b=9.81\cdot\mathrm{GCR}\cdot m_{\mathrm{robot}}$) with a fixed offset from
the neck to capture the restoring torque. The physical tendon--spring
knee mechanism is approximated without explicitly modeling the tendon
or four-bar linkage. Tendon actuation is represented by unilateral PD
control ($K_p=1.00$, $K_d=0.08$), reproducing its tension-only behavior,
with torque limited to $0.0847$~N$\cdot$m; an opposing linear torsional
spring models the passive spring. The neck and hip joints remain passive.
Designs vary in GCR, SPCF, and, when applicable, the femur and tibia lengths.

\noindent \textbf{Task and Policy description}. We consider two locomotion tasks: (i) obstacle traversal and (ii) ramp walking. In the obstacle task, a cuboidal block of width = 1m, length = 2m and variable height (h) is placed 0.5 m in front of the robot. The performance is evaluated by the maximum height (h) of obstacle crossed. We use a curriculum where the obstacle height increases/decreases by 1 cm if the robot succeeds/fails at the current level.
In ramp walking, a triangular ramp of width = 1.5m, length = 2m and variable height (h) is placed 0.5m in front of the robot. Performance on this task is primarily measured by the maximum slope grade (\(\gamma = \frac{\text{ramp height}}{\text{ramp base}} \times 100 \%\)) traversable by the robot. Here, we use a curriculum where the ramp height (h) increases/decreases by 2~cm for success/failure at the current difficulty level. We use the Tree-Parzen Estimator (TPE)~\cite{TPE_BO} algorithm in Optuna \cite{optuna_2019} for design optimization and Proximal Policy Optimization (PPO)~\cite{Schulman2017ProximalPO} for RL training.

All specialists in GES are MLPs of size \{128, 64, 32\}. We compare GES against -- \textbf{Oracle}: TPE-based design optimization in which a separate PPO policy is trained from scratch for every candidate morphology and two universal policy baselines -- \textbf{e2e MLP}: a large monolithic MLP of size \{512, 512, 512\}. \textbf{e2e MoE}: a soft-gated mixture-of-experts where each expert has size \{256, 128, 64\}. Both baseline policies are trained end-to-end, hence the prefix e2e. All policies are conditioned on the design, and trained with morphology randomization across $\mathcal{D}$. 
For a GES specialist, the morphology randomization bounds are dictated by its respective Gaussian distribution at the particular iteration. We use the complete design bounds in Tab.~\ref{tab:design_space} for morphology randomization in case of e2e MLP and e2e MoE. For GES and e2e MoE, we use K = 6 experts/specialists. For the amortized methods (GES, e2e MLP, and e2e MoE), we fix the total training budget to \(B=4000\) PPO iterations.
For GES, \(n_\text{PPO}=1500\) during the initial iteration and \(200\) thereafter. Across 2D, 3D and 4D, the number of GES iterations is at most seven. If GES reaches 95\% coverage before exhausting the budget \(B\), we fine-tune the specialists for the remaining PPO iterations on the converged design distributions.

\begin{figure}
    \centering
    \includegraphics[width=1\linewidth]{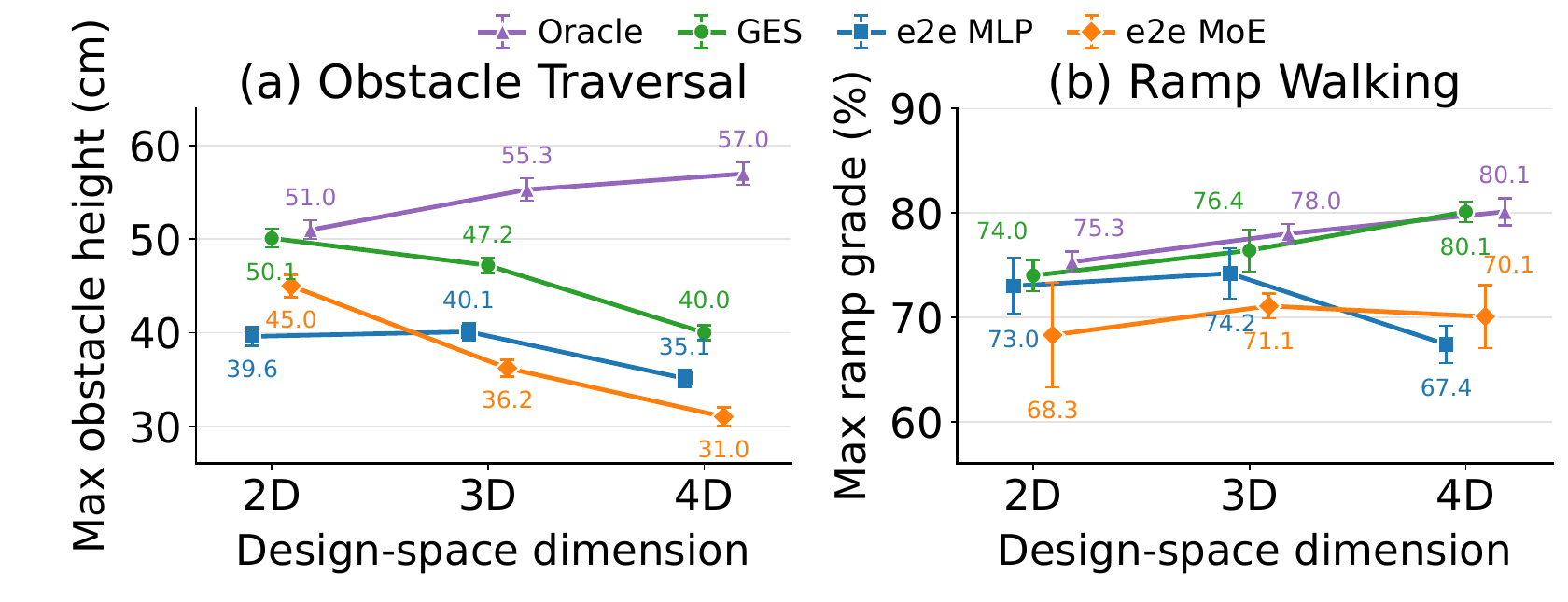}
    \caption{Effect of morphology-space dimensionality on the best design performance of Oracle, GES (ours), MLP, and MoE for (a) obstacle traversal and (b) ramp locomotion. Experiments repeated over 2 seeds of BO \(\times\) 2 seeds of PPO.}
    \label{fig:performance-vs-design-dim}
    \vspace{-1em}
\end{figure}

  
  

\subsection{GES Outperforms Universal-Policy Baselines}
\label{sec:main-results}


\noindent\textbf{Obstacle Traversal}. GES consistently finds higher performing designs than both universal-policy baselines across design-space dimensions (Fig. \ref{fig:performance-vs-design-dim}a). 
In the full 4D space, GES discovers a design that traverses a 40 cm obstacle, compared with 35.1 cm for e2e MLP and 31.0 cm for e2e MoE, although the Oracle reaches 57.0 cm. However, Oracle's higher performance comes at the cost of \(3\times\) higher compute requirement as discussed in Sec.~\ref{sec:compute_cost}. 
This advantage is consistent with GES providing stronger control across the morphology space: over 1000 uniformly sampled 4D designs, GES outperforms e2e MLP and e2e MoE on 89.8\% and 76.5\% of designs, respectively. 
Thus, the specialist population provides a stronger reusable evaluator than either universal-policy baseline, enabling TPE to identify higher-performing morphologies without per-design policy training. 

\noindent\textbf{Ramp Walking}. The same trend is stronger for ramp walking (Fig. \ref{fig:performance-vs-design-dim}b). GES discovers a design reaching an 80.1\% ramp grade compared with 67.4\% for e2e MLP and 70.1\% for e2e MoE, matching the Oracle in this setting. Across the 1000 uniformly sampled designs, GES outperforms e2e MLP and e2e MoE on 98.1\% and 94.9\% of designs, respectively. These results further support that improved control across morphology space translates into more effective design evaluation and, ultimately, better designs.

\begin{figure}[!t]
    \centering
    \includegraphics[width=0.6\linewidth]{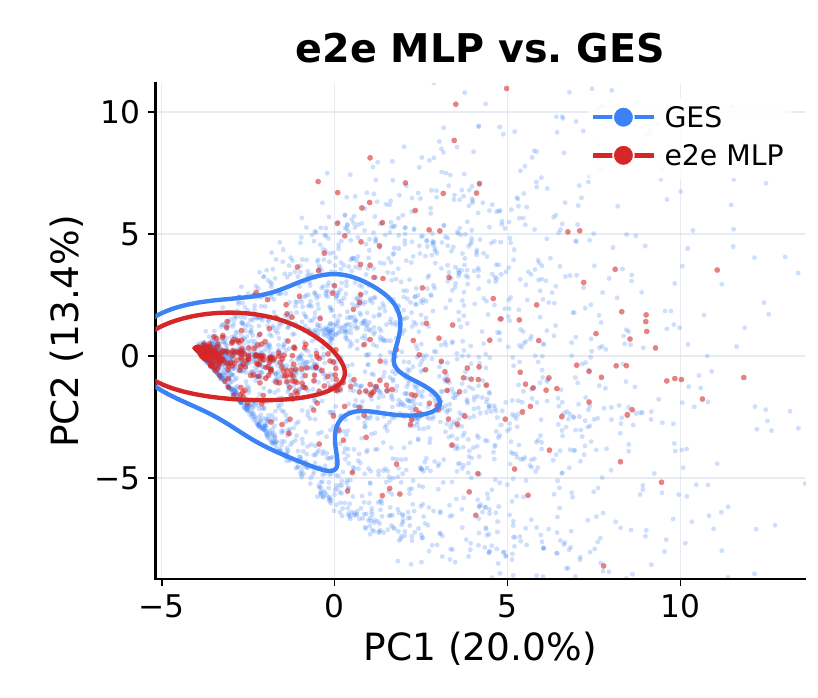}
    \caption{PCA visualization of toe-height behaviors for GES vs e2e MLP across the first two components preserving 33.4\% variance. Each point represents one rollout. Contours enclose 50\% of the highest density regions.} 
    \label{fig:behavioral-diversity}
    \vspace{-1em}
\end{figure}

\begin{figure}[t!]
    \centering
    \includegraphics[width=0.45\textwidth]{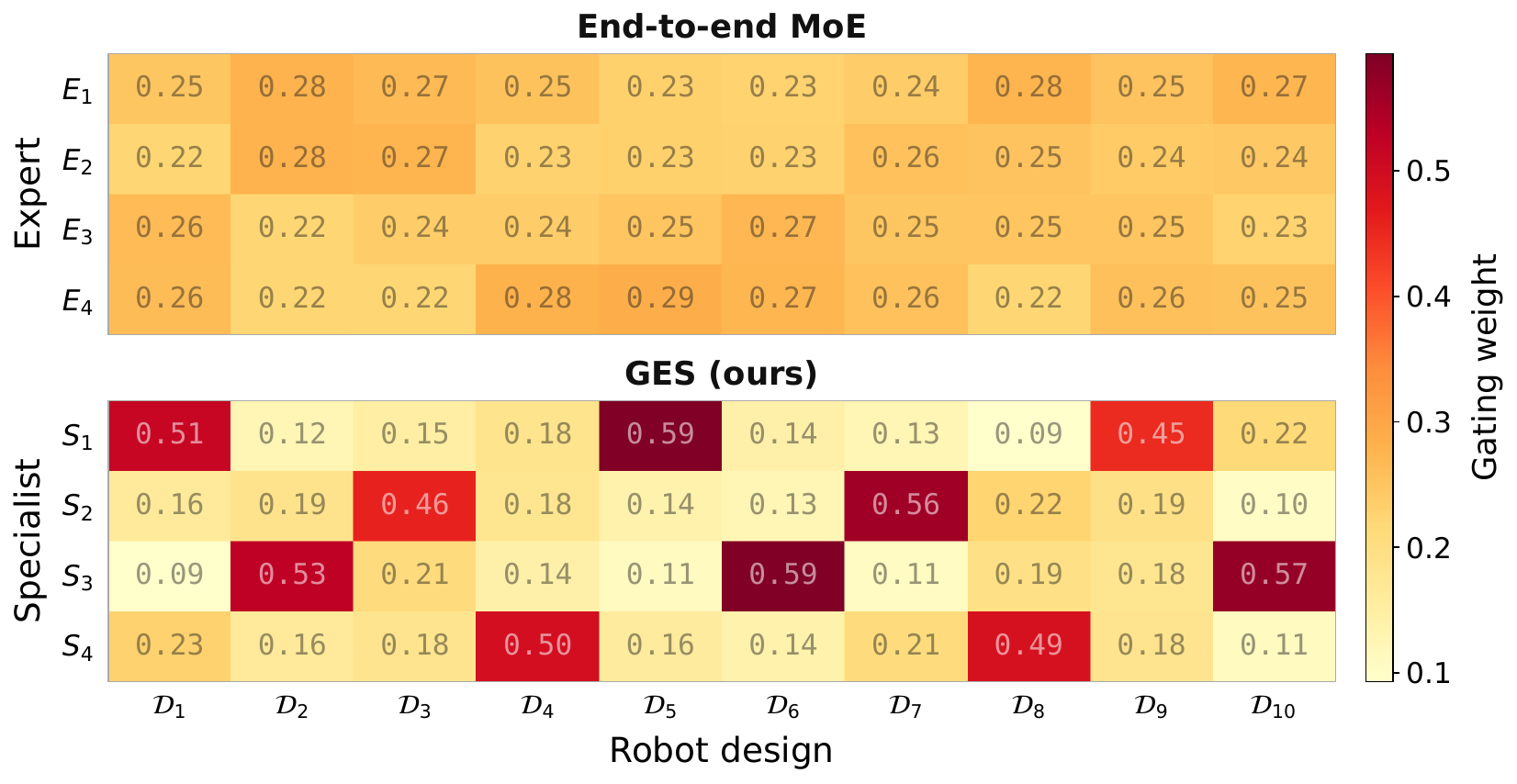}
    \caption{Expert contribution distribution for 4 experts across 10 designs drawn from distinct design-space regions. After convergence, the e2e MoE router distributes each design equally among all experts, whereas, 
    GES assigns each design to its dominant specialist.} 
    \label{fig:moe-vs-ges-heatmap}
    \vspace{-2em}
\end{figure}


\begin{figure*}[t]
    \centering
    \includegraphics[width=\textwidth]{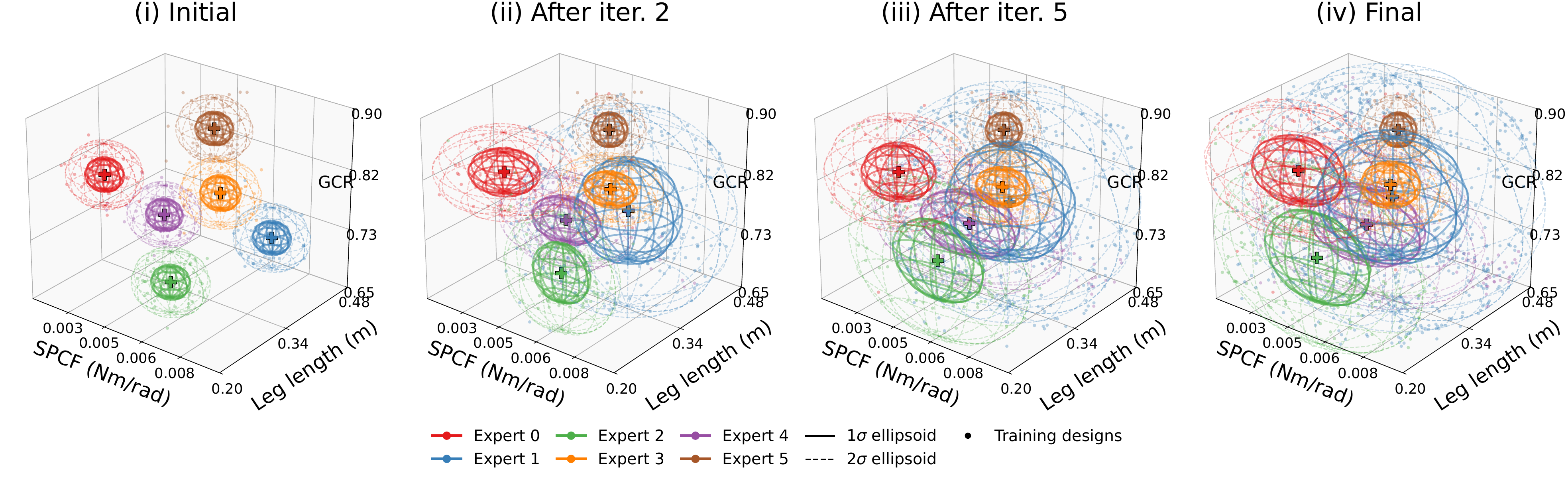}
    \caption{
    Evolution of six specialists (experts) in the 3D design space. For visualization only, we initialize coverage at \(\hat{c}_0=30\%\)}.
    \label{fig:state-evolution}
    \vspace{-1em}
\end{figure*}

\subsection{GES Preserves Behavioral Diversity}
\label{sec:behavior-analysis}

\noindent \textbf{Specialist Territory Evolution}.
Fig. \ref{fig:state-evolution} shows the Gaussian territory evolution of GES specialists in the 3D design space. Here, each point represents a unique design and a point-cloud denotes the design distribution of the color-coded specialist. The specialist distributions gradually grow and spread throughout iterations until they collectively tile the design space.


\noindent \textbf{GES vs e2e MLP}. We quantify behavioral diversity using toe trajectories from 500 uniformly sampled morphologies. Each rollout is represented by an 80-\textit{d} descriptor comprising the vertical positions of both toes over 40 timesteps. We fit Principal Component Analysis (PCA) on balanced samples from GES and e2e MLP and retain 15 components explaining 95.1\% of the variance. Fig.  \ref{fig:behavioral-diversity} visualizes the first two components, where GES spans behavioral regions beyond those covered by e2e MLP.

We measure this difference using bidirectional \(k\)-nearest-neighbor coverage in the 15-\textit{d} PCA space. 
For \(k=5\), GES covers 93.4\% of the e2e MLP behaviors (95\% CI over GES resampling: \(88.6\text{--}96.8\%\)), whereas the e2e MLP covers only 72.8\% of GES behaviors (95\% CI over GES resampling: \(69.4\text{--}76.2\%\)); this asymmetry holds for \(k\in\{3,5,10\}\). This indicates that GES retains most  behaviors available to the monolithic policy while additionally representing behaviors absent from it.

\noindent\textbf{GES vs e2e MoE}. We next examine whether the MoE baseline develops morphology-dependent specialization. Fig. \ref{fig:moe-vs-ges-heatmap} shows that its router assigns nearly uniform weights across experts over the design space, indicating limited expert differentiation. In contrast, GES develops distinct specialist territories with dominant assignments across different morphology regions. Thus, increasing policy capacity through an MoE architecture alone does not recover the specialization induced by GES.


\subsection{GES Balances Co-design Performance and Efficiency}
\label{sec:compute_cost}

GES achieves substantially higher design performance than the universal policy baselines at comparable co-design cost, while remaining significantly faster than the Oracle. GES even matches the Oracle's performance in Ramp Walking. 
We compare all methods under the same budget of \(T=200\) TPE design evaluations on a single NVIDIA RTX 4090 GPU. The total wall-clock costs are approximately 25.7 h, 26.7 h, and 27.7 h for e2e MLP, e2e MoE, and GES, respectively, compared with 85.0 h for the Oracle. Thus, GES incurs only a modest computational overhead over the universal-policy baselines while providing consistently stronger design performance (Fig. \ref{fig:performance-vs-design-dim}), and reduces total co-design time by \textbf{67.4}\% relative to the Oracle.

The computational savings of all three amortized methods arise from avoiding retraining for every candidate design. The Oracle trains a new policy for 1,500 PPO iterations (\(\sim\) 20 min) and evaluates it (\(\sim\) 5 min) for every candidate. In contrast, e2e MLP and e2e MoE incur upfront training costs of approximately 4h and 5h, respectively, after which candidates are evaluated zero-shot in \(\sim\) 6 min. GES incurs a larger upfront cost of 9.4 h to complete seven Train-Probe-Refit iterations for \(K=6\) specialists. Tab. \ref{tab:compute_cost} summarizes the resulting end-to-end computational costs.

These results highlight the performance-efficiency trade-off between the approaches. The Oracle provides the strongest per-design controller, but at substantially higher computational cost. The e2e MLP and e2e MoE achieve the lowest co-design times but find lower-performing designs, while GES improves design performance at only a small additional cost. All timings report single-GPU wall-clock time; although GES specialists can be trained and probed independently across multiple GPUs, we exclude this potential parallel speedup from the reported savings.


\subsection{GES Performance Improves with Specialist Capacity}
\label{sec:ablations}
\begin{figure}
    \centering
    \includegraphics[width=0.5\linewidth]{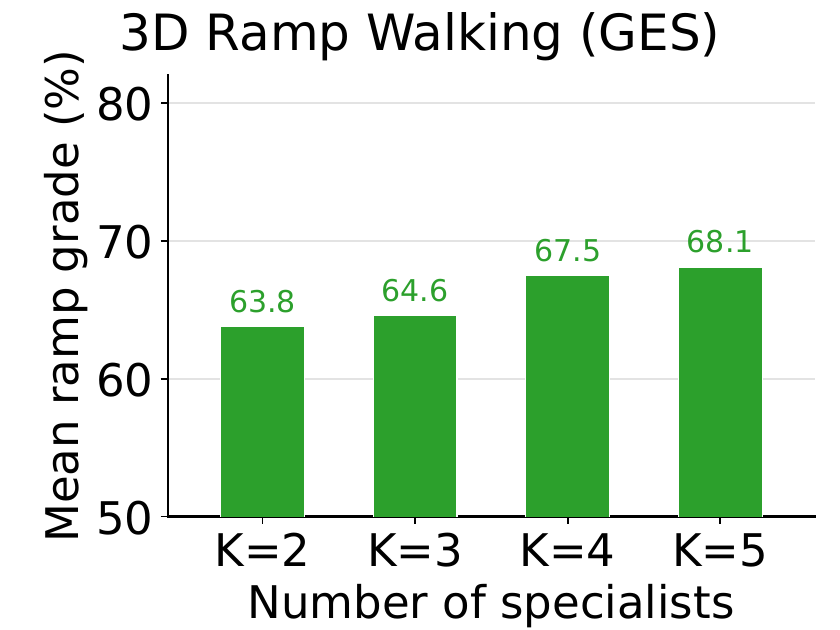}
    \caption{GES performance sensitivity to \# specialists in ramp walking in the 3D design space.}
    \label{fig:k-sensitivity-ges}
    \vspace{-1em}
\end{figure}

We analyze the impact of the number of specialists \(K\) on the final performance of GES. We vary \(K \in \{2,3,4,5\}\) for ramp walking in the 3D design space. The optimized ramp grade increases from 63.8\% at K=2 to 68.1\% at K=5, with gains diminishing after K=4 (Fig.~\ref{fig:k-sensitivity-ges}).
This suggests that additional specialists improve design-space coverage up to a moderate capacity, after which the benefit begins to saturate.

\begin{table}[t]
    \centering
    \caption{Single-GPU (parallel environments) computational cost for Oracle and GES. Both methods use the same $T=200$ TPE evaluation budget.} 
    \label{tab:compute_cost}
    \resizebox{\columnwidth}{!}{%
    \begin{tabular}{lcc}
        \toprule
        \textbf{Computational component}
        & \textbf{Oracle}
        & \textbf{GES} \\
        \midrule
        TPE design evaluations               & 200 & 200 \\
        PPO training / candidate             & 1,500 iter. ($\sim$20 min) & -- \\
        Evaluation / candidate               & $\sim$5 min & $\sim$5 min \\
        TPE overhead / candidate             & $\sim$0.5 min & $\sim$0.5 min \\
        \midrule
        200-evaluation design search         & 85.0 h & 18.3 h \\
        Specialist PPO training              & -- & 5.33 h \\
        Specialist probing                   & -- & 3.50 h \\
        Candidate/probe generation           & -- & 0.41 h \\
        Gaussian refit + initialization      & -- & 0.13 h \\
        \midrule
        \textbf{Upfront GES cost}            & -- & \textbf{9.4 h} \\
        \textbf{Total co-design time}        & \textbf{85.0 h} & \textbf{27.7 h} \\
        \textbf{Reduction}                   & -- & \textbf{67.4\%} \\
        \bottomrule
    \end{tabular}%
    }
\end{table}



\subsection{Optimized BALLU designs differ across tasks}
Tab.~\ref{tab:design_space} summarizes the optimized design for the two tasks in comparison to the baseline design. Fig.~\ref{fig:sim-motion-strip} shows a motion strip comparing the optimized and baseline designs in ramp walking.

\noindent\textbf{Design Analysis.} In obstacle traversal, higher GCR and SPCF consistently produce better performance: high buoyancy reduces effective weight to enable flight and high spring stiffness stores the potential energy needed for jumping. Longer legs also help but show a weaker correlation with performance. The optimized design is $\texttt{GCR}=0.88$, $\texttt{SPCF}=0.008$~Nm/rad, \(l_f = 45.58\)~cm, \(l_t = 45.58\)~cm whose limb dimensions are approximately the same as baseline design.

In ramp walking, we find kinematics to play a much stronger role. Short-legged robots tend to perform better than their long-legged counterparts. We attribute this to two factors: (a) Short legs provide better mobility on a slope. Robots often turn during the ascent, eventually slipping down and failing the task. The turning tendency increases with longer legs. (b) effective reduction of CoM height. Lower \(z_{COM}\) leads to less counter torque from gravity which helps the robot maintain balance on steep slopes. The resulting behavior qualitatively resembles human uphill walking, characterized by a bent posture and increased knee flexion \cite{walking_posture_while_climbing}. The optimized design is $\texttt{GCR}=0.70$, $\texttt{SPCF}=0.005$~Nm/rad, $l_f = 30$~cm, $l_t = 28$~cm.



\begin{figure}[t]
    \centering
    \includegraphics[width=0.48\textwidth]{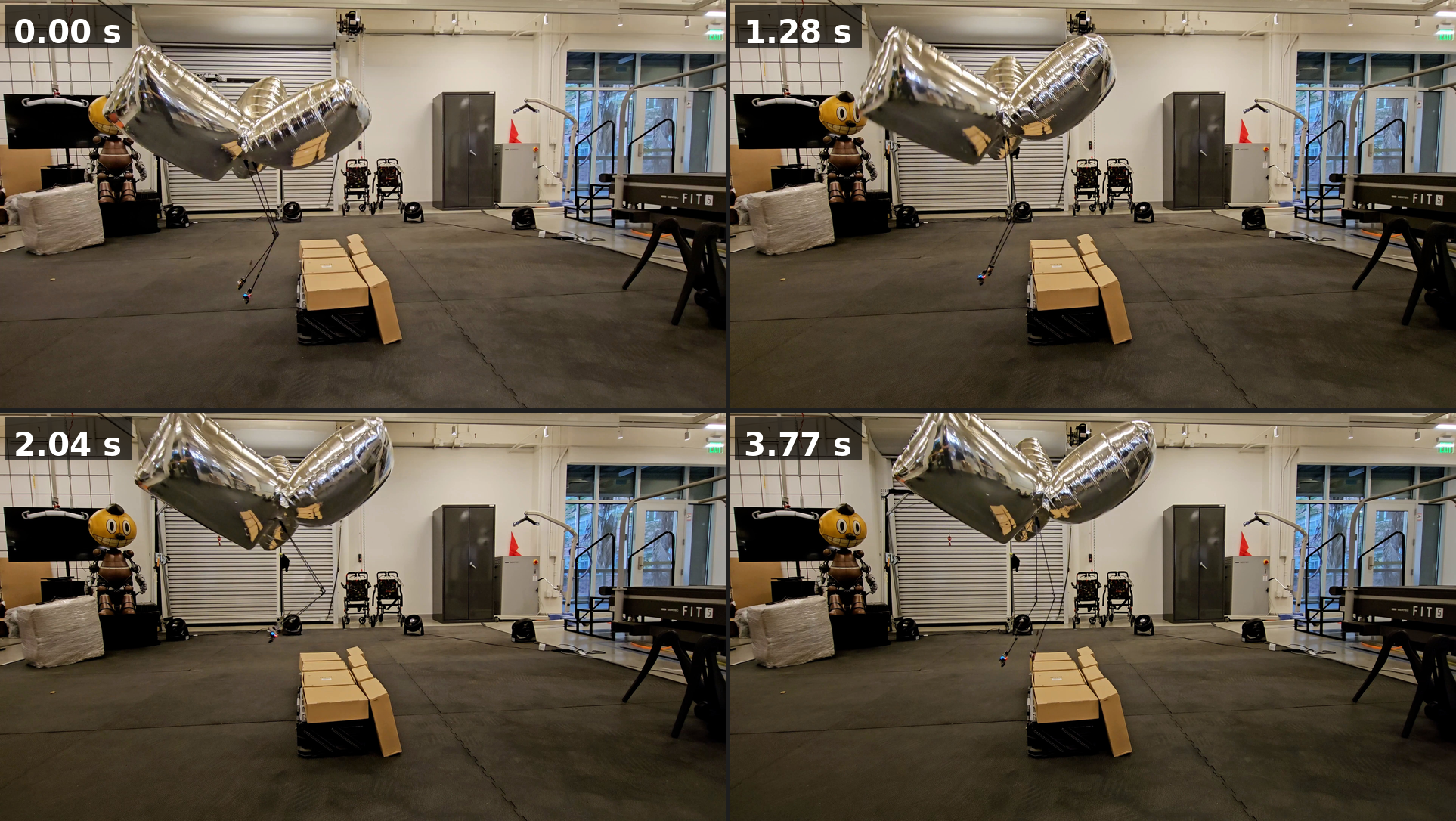}
    \caption{Optimized BALLU ($\text{GCR}=0.88$, $\text{SPCF}=0.008$~Nm/rad) teleoperated to traverse a 24~cm obstacle, while baseline design traverses upto 8~cm.}
    \label{fig:hardware_composite}
    \vspace{-1em}
\end{figure}

\noindent\textbf{Hardware Results.} We additionally evaluate the morphology identified by the proposed framework on the physical BALLU platform. Since this work focuses on efficient morphology optimization rather than sim-to-real transfer, we evaluate the optimized and original morphologies using \textit{teleoperated control}, separating morphology-level evaluation from the challenge of transferring a learned policy to hardware. Under the same teleoperation setting, the optimized morphology successfully traverses obstacles up to 24 cm, compared with 8 cm for the original design (Fig.~\ref{fig:hardware_composite}). These results provide physical evidence that the morphology identified through simulation improves obstacle-traversal capability under teleoperated control. We emphasize that this experiment evaluates the \textit{optimized morphology}, rather than the performance of a GES specialist policy on hardware. 

\section{CONCLUSIONS}

We presented GES, an RL-based co-design framework that amortizes controller learning across candidate designs while preserving specialization across the morphology space. 
Simulation results showed the advantages of GES over existing universal policy baselines. The teleoperated hardware experiment on the optimized design further supports the usefulness of our framework.

In the future, we plan to deploy the learned specialist policies on hardware to evaluate the complete co-design pipeline.
GES currently assumes a fixed number of specialists and continuous, bounded design parameters. Future work can investigate adaptive specialist allocation, more general design spaces, and application to other robotic platforms. 

\section*{ACKNOWLEDGMENT}
We used Claude to assist with experimental code, grammar correction, and minor editing. All AI-generated content was reviewed and validated by the authors.




\bibliography{IEEEabrv,refs}

\end{document}